\documentclass{ecai}

\usepackage{latexsym}
\usepackage{amssymb}
\usepackage{amsmath}
\usepackage{amsthm}
\usepackage{booktabs}
\usepackage{enumitem}
\usepackage{graphicx}
\usepackage{color}

\usepackage{hyperref}
\usepackage{multicol}
\usepackage{multirow}
\usepackage{graphics}
\usepackage{makecell}
\usepackage{xcolor}
\usepackage{subfigure}
\usepackage{wrapfig}
\usepackage{bm}
\usepackage{stfloats}
\usepackage{diagbox}

\newcommand{\BibTeX}{B\kern-.05em{\sc i\kern-.025em b}\kern-.08em\TeX}
\newcommand{\sep}[1]{\scriptsize\textcolor{gray}{#1}}

\begin{document}

\begin{frontmatter}

\paperid{123}

\title{Context Enhancement with Reconstruction as Sequence \\ for  Unified Unsupervised Anomaly Detection}

\author[A,B,C]{\fnms{Hui-Yue}~\snm{Yang}}
\author[A,B]{\fnms{Hui}~\snm{Chen}}
\author[A,B]{\fnms{Lihao}~\snm{Liu}}
\author[A]{\fnms{Zijia}~\snm{Lin}}
\author[A,B]{\fnms{Kai}~\snm{Chen}} 
\author[D]{\fnms{Liejun}~\snm{Wang}}
\author[A,B]{\fnms{Jungong}~\snm{Han}}
\author[A,B]{\fnms{Guiguang}~\snm{Ding}\thanks{Corresponding Author. Email: dinggg@tsinghua.edu.cn.}}

\address[A]{Tsinghua University}
\address[B]{BNRist}
\address[C]{Hangzhou Zhuoxi Institute of Brain and Intelligence}
\address[D]{Xinjiang University}

\begin{abstract}
Unsupervised anomaly detection (AD) aims to train robust detection models using only normal samples, while can generalize well to unseen anomalies. Recent research focuses on a unified unsupervised AD setting in which only one model is trained for all classes, \textit{i.e.,} n-class-one-model paradigm. Feature-reconstruction-based methods achieve state-of-the-art performance in this scenario. However, existing methods often suffer from a lack of sufficient contextual awareness, thereby compromising the quality of the reconstruction. To address this issue, we introduce a novel Reconstruction as Sequence (RAS) method, which enhances the contextual correspondence during feature reconstruction from a sequence modeling perspective. In particular, based on the transformer technique, we integrate a specialized RASFormer block into RAS. This block enables the capture of spatial relationships among different image regions and enhances sequential dependencies throughout the reconstruction process. By incorporating the RASFormer block, our RAS method achieves superior contextual awareness capabilities, leading to remarkable performance. Experimental results show that our RAS significantly outperforms competing methods, well demonstrating the effectiveness and superiority of our method. Our code is available at \href{https://github.com/Nothingtolose9979/RAS}{https://github.com/Nothingtolose9979/RAS} 
\end{abstract}

\end{frontmatter}

%%%%%%%%%%%%%%%%%%%%%%%%%% Introduction %%%%%%%%%%%%%%%%%%%%%%%%%%%%%%%%%%

\section{Introduction}

\begin{figure}[ht]
\centering
\includegraphics[width=\linewidth]{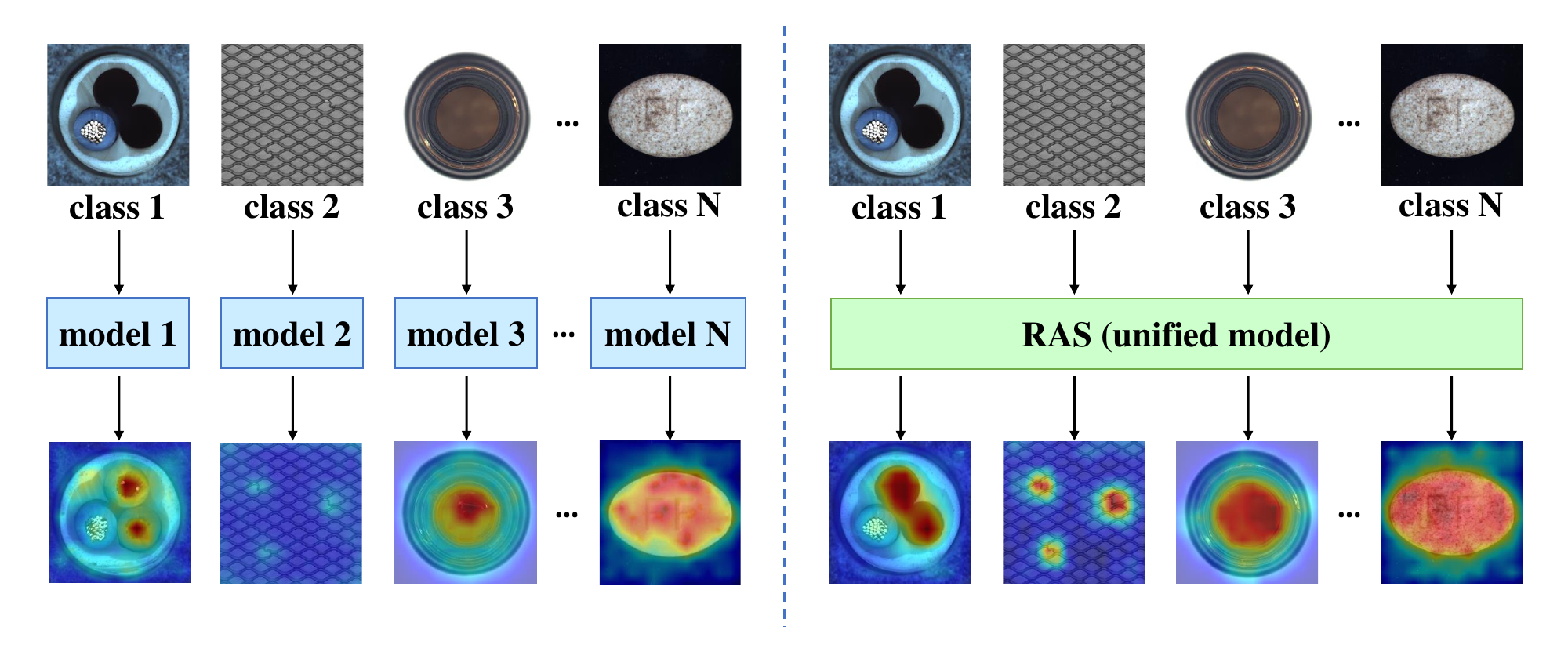}
% \vspace{-20pt}
\caption{Comparison of different paradigms in anomaly detection. \textbf{Left:} n-class-n-model paradigm, where \textbf{separate} models are trained for each class. \textbf{Right:} n-class-one-model paradigm, utilizing a \textbf{unified} model to detect anomalies across all classes.} 
\label{fig:paradigm}
\vspace{20pt}
\end{figure}

Anomaly detection (AD) aims to identify outliers or abnormal regions for an input image. It is widely used in various fields such as industrial manufacturing \cite{mvtec,visa,destseg,simplenet}, healthcare \cite{medical,diffusionmed,med1}, surveillance \cite{hypervd,mgfn,surveillance1} and fraud detection\cite{fraud1,fraud2,fraud3,fraud4}. Developing optimal AD models is challenging due to the rarity of anomalies in real-world scenarios. Researchers have explored unsupervised anomaly detection without requiring anomaly-specific data. Nonetheless, they often build \textbf{separate} models for each class, \textit{i.e.,} the n-class-n-model paradigm shown in Fig.~\ref{fig:paradigm} (left). However, due to the diversity of anomaly classes, such a paradigm may not be the best solution, especially as the number of classes increases~\cite{omnial}. 

Recently, developing a robust AD framework that achieves \textbf{unified} unsupervised anomaly detection has gained much attention~\cite{uniad,omnial}. Such a unified setting can detect different anomalies for all classes with only one AD model, \textit{i.e.,} n-class-one-model paradigm shown in  Fig.~\ref{fig:paradigm} (right). In this scenario, feature-reconstruction-based method has emerged as a popular method, owing to its simplicity, impressive detection performance, and robustness. These techniques focus on reconstructing the visual features of an input image during the feature reconstruction process, with anomaly regions identified by comparing the original image feature and the reconstructed one. For example, UniAD \cite{uniad} first incorporates the transformer architecture with feature jittering and neighbor masked attention to amplify feature differences and improve the accuracy of anomaly detection. UniCon-HA \cite{wang2023unilaterally} proposes unilaterally aggregated contrastive learning to obtain the concentrated inlier distribution as well as the dispersive outlier distribution.

\begin{figure}[ht]
\centering
\includegraphics[width=\linewidth]{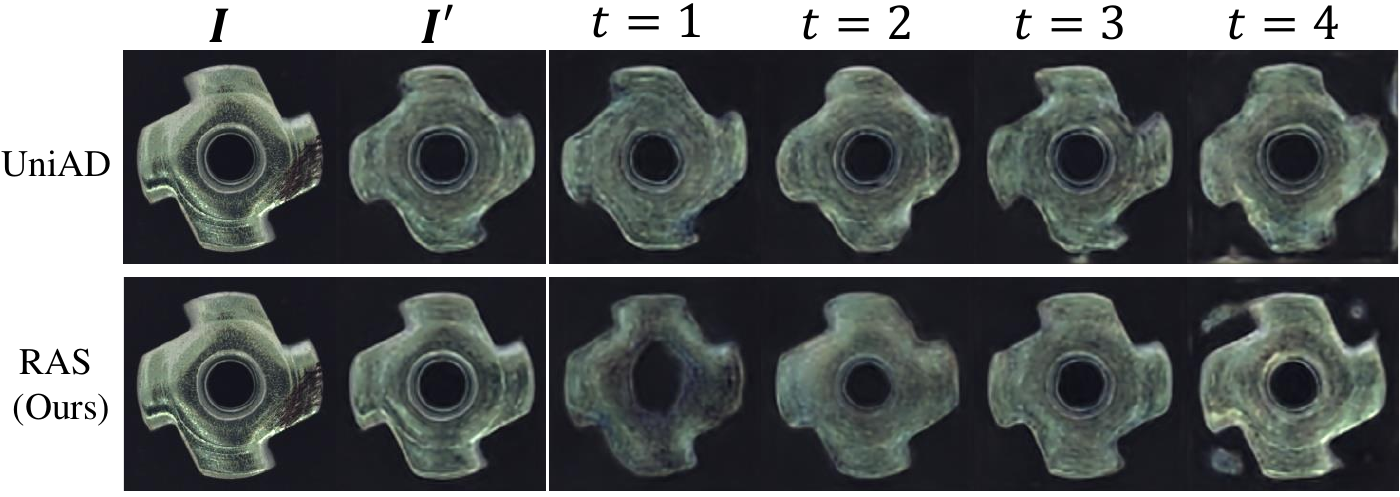}
\caption{\textbf{Top:} inspection of the reconstruction failure of UniAD. \textbf{Bottom:} illustration of the superior reconstruction quality of our proposed RAS method. $\bm{I}$ is an anomalous metal nut.} 
\label{fig:problem}
\vspace{20pt}
\end{figure}

Despite the promising results, the effectiveness of feature-reconstruction-based methods heavily relies on the quality of the reconstructed features, which is difficult to achieve. To more intuitively demonstrate this issue, we map the image features back into the RGB image using an image decoder\footnote{The image decoder is designed to over-fit the test distribution, enabling it to perfectly map the image features back into the RGB image and reflect the quality of the reconstructed features.}. The alignment between the original image and the reconstructed image determines how well the feature reconstruction captures visual differences. However, as shown in Fig.~\ref{fig:problem}, we observe that UniAD (used as a representative method) fails to adequately capture crucial object details, such as edges and lighting. This limitation may lead to false positive predictions in anomaly detection and ultimately result in inferior performance.

Furthermore, we further investigate the feature reconstruction process step-by-step for a deeper understanding. We encourage each reconstructed feature, \textit{i.e.,} the output of each decoder, to be mapped back into the RGB image space. By inspecting the differences between these reconstructed images, as shown in Fig.~\ref{fig:problem}, we observe that there are minimal variations among the successively reconstructed images in UniAD. This indicates that the decoder fails to capture the intricate patterns already reconstructed by the preceding decoder, leading to limited contextual awareness throughout the reconstruction process. This naturally raises the question: \textit{How to improve contextual correspondence during the feature reconstruction to enhance anomaly detection?}

In this paper, we answer this question through a novel \textbf{R}econstruction \textbf{A}s \textbf{S}equence (\textbf{RAS}) method, which rethinks the feature reconstruction process from the perspective of sequence modeling for the unified unsupervised anomaly detection. Specifically, we consider each decoder layer as one step in the sequence model. In this sense, we expect that \textit{sequential dynamics} within different steps and \textit{spatial dynamics} in the visual context can be simultaneously captured for the feature reconstruction. Consequently, we derive a RASFormer block that adapts the transformer architecture with a novel strategy of adaptive gating to enhance the contextual awareness ability. Benefiting from the gating strategy, our RAS method can comprehensively learn the sequential dynamics during feature reconstruction. Besides, the spatial discrepancies among the visual regions can be well grasped and enhanced for anomaly detection. As a result, our RAS can achieve superior reconstruction quality (see Fig.~\ref{fig:problem}) and anomaly detection performance.

Overall, our contributions are three folds:
\begin{itemize}
\item We thoroughly consider the contextual awareness capability during the feature reconstruction for the unified unsupervised AD. A novel Reconstruction as Sequence (RAS) method is proposed, which rethinks the feature reconstruction process from the sequence perspective. 
\item We introduce a generic RASFormer block to effectively enhance the contextual correspondence during the feature reconstruction, resulting in remarkable reconstruction outcomes.
\item Experimental results on several benchmark datasets show that the proposed RAS can achieve state-of-the-art performance, well demonstrating the effectiveness and superiority of the proposed method.
\end{itemize}

%%%%%%%%%%%%%%%%%%%%%%%%%% Related Work %%%%%%%%%%%%%%%%%%%%%%%%%%%%%%%%%%
\section{Related Work}
\textbf{Unsupervised anomaly detection.} Due to the limited availability of anomalous samples, unsupervised learning methods are commonly employed for anomaly detection in real scenarios, \textit{e.g.,} industrial quality inspection.
Early works incorporate patch-level embedding~\cite{psvdd}, geometric transformation~\cite{geotrans}, and elastic weight consolidation~\cite{panda}, resulting in great improvement. 
Some works use a pre-trained backbone to extract features and model the normal distribution~\cite{padim,modelingdistribution}, followed by a distance metric to identify anomalies.
Nonetheless, these methods are computationally expensive due to the need of memorizing all image features, making them impractical when facing a large number of images.
Knowledge distillation methods \cite{us,mkd,destseg} distinguish the difference between teacher and student for anomaly detection.

Reconstruction-based works assume that reconstruction models trained solely on normal samples perform well in normal regions but fail in anomalous regions~\cite{l2ae_ssimae,utrad,vevae}. Representative works include using generative networks \cite{aescstrain,ganomaly,vevae}, pseudo-anomaly \cite{gtod,stainshapednoise}, and synthesizing anomalies on normal images \cite{draem,cutpaste}. While these methods have shown success in separate one-class-one-model anomaly detection (AD) scenarios, their performance tends to be subpar in the unified n-class-one-model scenario~\cite{uniad,omnial}.

\textbf{Unified unsupervised AD.} Conventional AD methods require training separate models for each class, which becomes costly as the number of classes increases. Recently, constructing a unified model for multi-class anomaly detection has gained popularity in the research community. RegAD~\cite{regad} addresses few-shot anomaly detection by training a single generalizable model, utilizing a limited number of normal images for each category during training. UniAD~\cite{uniad}  employs a feature-reconstruction approach to pinpoint anomalous regions with the transformer architecture. OmniAL~\cite{omnial} presents a panel-guided method to synthesize anomalies and achieve image reconstruction using dilated channel and attention mechanism \cite{imram,attention}. These works primarily concentrate on capturing discriminative patterns that can identify anomalies by misaligning them with the normal distribution. Our RAS essentially shares a similar objective but takes it a step further by emphasizing context enhancement from a novel sequence modeling perspective \cite{grn,bert}. We show that our RAS can obtain remarkable reconstruction quality and thus achieve superior performance for unified unsupervised anomaly detection.

%%%%%%%%%%%%%%%%%%%%%%%%%% Method %%%%%%%%%%%%%%%%%%%%%%%%%%%%%%%%%%

\begin{figure*}[ht]
\centering
\includegraphics[width=0.9\linewidth]{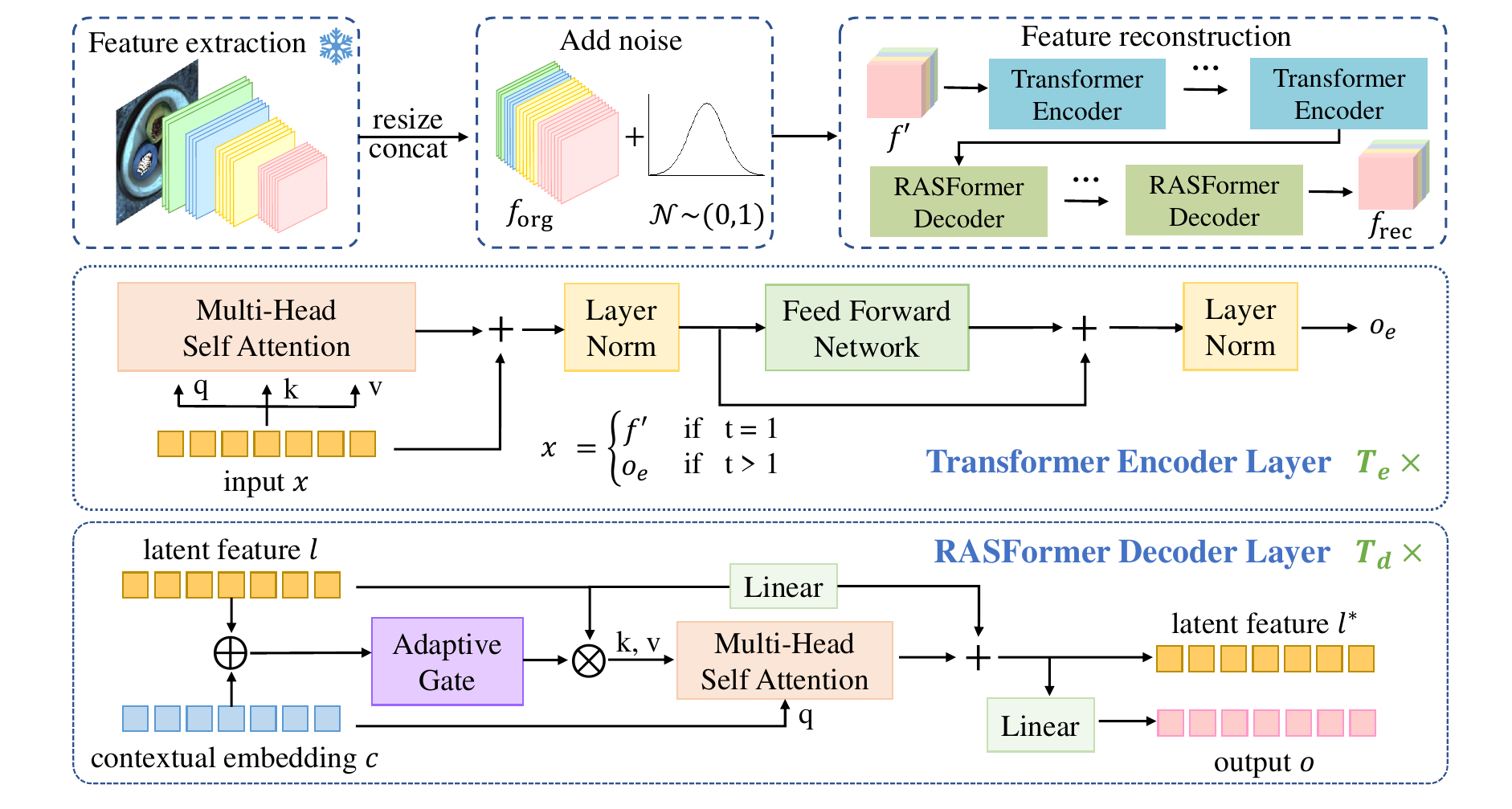}

\caption{Overview of the proposed RAS framework for the unified unsupervised anomaly detection. We enhance the contextual awareness capability during feature reconstruction via a specially designed RASFormer block.
The uppermost panel depicts the pipeline, while the two boxes below illustrate the detailed architecture of encoder-decoder to perform feature reconstruction.
} 
\label{fig:res_mvtec}
\vspace{10pt}
\end{figure*}

\section{Reconstruction as Sequence (RAS)}
\label{sec:ras}

\subsection{Preliminary}
\label{sec:pre}

\textbf{Image feature extraction.} In the feature-reconstruction-based model, the goal is to align the reconstructed feature $\bm{f}_{\rm rec}$ with the original image feature $\bm{f}_{\rm org}$. To accomplish this, we employ a pre-trained convolutional neural network (CNN) \cite{resnet,efficientnet,repvit} as the backbone for extracting the original image feature. This backbone is denoted as $\phi$, and the process of deriving features from the image $\bm{I}$ can be represented as 
$\phi(\bm{I})=\{\bm{f}_1,...,\bm{f}_n\}$, where $n$ is the number of feature levels. Consequently, for each feature level, we apply a 3$\times$3 average pooling operation, resize them to the same size, and concatenate all the features along the channel dimension, yielding a comprehensive feature map:
\begin{equation}
\label{equ:f_origin}
    \bm{f}_{\rm org} \in \mathbb{R}^{C_{\rm org}\times (H \times W)} = {\rm concat} \{\bm{f}_k|k=1,..,n\}
\end{equation}
where $C_{\rm org}$, $H$, and $W$ are the feature dimension, height and width of the feature map, respectively.

\textbf{Transformer layer.} Transformer~\cite{attention,bert} has emerged as a foundational architecture in the field of computer vision \cite{resnet,vit,yolov10,swin}. A transformer layer comprises two essential sub-layers: the multi-head self-attention (MHSA) and the feed-forward network (FFN). To enhance training efficiency and performance, residual connections~\cite{resnet} and layer normalization (LN)~\cite{layernorm} are applied to each sub-layer independently. Here, we utilize a post-LN transformer architecture~\cite{onlayernorm} to construct the transformer layer:
\vspace{-2pt}
\begin{equation}
    \resizebox{.99\linewidth}{!}{${\rm Transformer}(\bm{x}_q, \bm{x}_k, \bm{x}_v) = {\rm LN}({\rm FFN}({\rm LN}({\rm MHSA}(\bm{W}_q\bm{x}_q,\bm{W}_k\bm{x}_k,\bm{W}_v\bm{x}_v))))$}.
\end{equation}
where $\bm{x}_q$, $\bm{x}_k$, and $\bm{x}_v$ are input token sequences. $\bm{W}_q$, $\bm{W}_k$, and $\bm{W}_v$ are all learnable parameters. For ease of description, we omit the residual connection in the above equation.

\subsection{Feature Reconstruction from the Sequence Perspective}
\label{subsec:feature_recon}

\textbf{Denoised encoding.} The proposed RAS framework employs an encoder-decoder structure to reconstruct the image feature, \textit{i.e.,} $\bm{f}_{\rm org}$, which is derived by a CNN backbone, as depicted in Eq.~\ref{equ:f_origin}. 

We add noise to normal features and feed them into a transformer-based encoder, achieving a robust AD model to distinguish anomalies:
\begin{equation}
    \bm{f}' \in \mathbb{R}^{C_{\rm rec} \times (H \times W)} = \bm{W}_f(\bm{f}_{\rm org} + \bm{\epsilon})
\end{equation}
\vspace{-15pt}
\begin{equation}
    \resizebox{.99\linewidth}{!}{$\bm{o}_e \in \mathbb{R}^{C_{\rm rec} \times( H \times W)} = {\rm Transformer}_{T_e}(...{\rm Transformer}_1(\bm{f}', \bm{f}', \bm{f}'))$}.
    \label{equ:encoder}
\end{equation}
where $\bm{W}_f \in \mathbb{R}^{C_{\rm rec} \times C_{\rm org}}$ and $C_{\rm rec}$ is the dimension of the latent reconstruction space. $T_e$ is the number of encoders. $\bm{\epsilon}=\{\bm{\epsilon}^i, i\in[0,H \times W)\}$ are the noisy features added to $\bm{f}_{\rm org}$ during training, allowing the model to learn features of normal images through denoising:
\begin{equation}
    \bm{\epsilon}^i \sim N(\mu=0, \sigma^2=(\alpha \frac{||\bm{f}_{\rm org}^i||_2}{C_{\rm org}})^2)
    \label{eq:noise}
\end{equation}
where $\bm{f}_{\rm org}^i \in \mathbb{R}^{C_{\rm org}}$ is one element in $\bm{f}_{\rm org}$. $\alpha$ is the noise intensity to control the degree of noise. During the test phase, $\bm{\epsilon}$ is not applied.

\textbf{Sequence decoding.} UniAD~\cite{uniad} adopts conventional transformer layers to construct the decoder for feature reconstruction. Nonetheless, it is constrained in effectively capturing the contextual correspondence among decoding layers (see Fig. \ref{fig:problem}). In contrast, our proposed RAS framework can enhance the contextual correspondence by considering the feature reconstruction process from a sequence perspective. Specifically, at each decoding step of $t$, we are given the previously latent features $\bm{l}_{t-1}$. Next, we equip an individual context embedding $\bm{c}_t \in \mathbb{R}^{C_{\rm rec} \times (H \times W)}$ for each decoding step. A decoder $\theta_{\rm dec}^t$ consumes $\bm{l}_{t-1}$ and $\bm{c}_t$ and performs the mapping from the latent space to the image feature space:
\begin{equation}
\bm{l}_t\in \mathbb{R}^{C_{\rm rec} \times (H \times W)}, \bm{o}_t\in \mathbb{R}^{C_{\rm org} \times (H \times W)} = \theta_{\rm dec}^t(\bm{c}_t, \bm{l}_{t-1})
\label{equ:sequence}
\end{equation}
where $\bm{l}_t$ is the updated latent feature and $\bm{o}_t$ is the reconstructed feature. The decoding process in Eq.~\ref{equ:sequence} can be repeated several times, resulting in a sequence of reconstructions. We initialize the first latent feature with the output of the last encoder layer in Eq.~\ref{equ:encoder}, \textit{i.e.,} $\bm{l}_{0}=\bm{o}_e$. The final reconstructed feature $\bm{f}_{\rm rec}$ can be denoted as $\bm{f}_{\rm rec} = \bm{o}_{T_d}$, where $T_d$ is the number of decoders. 

It is worth noting that like the object query in DETR~\cite{detr}, after being well trained, the contextual embedding in the decoding step of $t$, \textit{i.e.,} $\bm{c}_t$, can be considered as the token query. This query provides a contextual prior assumption about $t$-th reconstructed features, \textit{i.e.,} $\bm{o}_{t}$. During the sequence of feature reconstruction, the latent features $\bm{l}_t$ and $\bm{l}_{t-1}$ are responsible for memorization of reconstruction knowledge. Therefore, the decoder $\theta_{\rm dec}^t$ should be powerful enough in the capability of context awareness among sequences, so that different knowledge can be uniformly captured in different decoding steps for better feature reconstruction. In light of this, we design a novel RASFormer block as the fundamental building block for decoders $\theta_{\rm dec}^t$. For ease of understanding, here we briefly represent the RASFormer block as a function, \textit{i.e.,} $\theta_{\rm dec}^t = {\rm RASFormer}_t(\cdot)$.

\subsection{RASFormer Block}
\label{sec:rasformer}
The RASFormer block serves as a fundamental module in the decoder, playing a crucial role in capturing contextual correspondence within the sequential feature reconstruction process. We adhere to two guiding principles when designing the RASFormer block: 1) \textit{sequential dynamics}, ensuring the awareness of the previously captured information, alleviating the need to readdress it in subsequent reconstruction processes; 2) \textit{spatial dynamics}, enabling the association between elements in the input context embedding $\bm{c}_t$ and those in the previous knowledge $\bm{l}_{t-1}$. To achieve this, we introduce a novel strategy of adaptive gating with transformers.

Specifically, given the prior knowledge, \textit{i.e.,} the previous latent reconstructed feature $\bm{l}$ and the current input context embedding $\bm{c}$ (for ease of description, we leave out the subscript $t$), we design an adaptive gate $A$ to filter the prior knowledge as follows:

\begin{align}
    \bm{a} &= A(\bm{l}, \bm{c})=\sigma(\bm{W}_A (\bm{l} \oplus \bm{c}))\\
    \bm{l}_{\rm A} &= \bm{a} \otimes \bm{l}
    \label{equ:reset}
\end{align}
where $\oplus$ is the concatenation of two tensors along the channel dimension. $\otimes$ represents the element-wise multiplication between two matrices. $\bm{W}_A$ is a learned weight matrix. $\sigma$ is an activation function (\textit{e.g.,} sigmoid). Note that all output tensors have the same shape as $\bm{l}$, \textit{i.e.,} $\mathbb{R}^{C_{\rm rec} \times (H \times W)}$. With the adaptive gate $A$, the current latent feature $\bm{l}_{\rm A}$ can adaptively retain relevant contextual information while disregarding unimportant details. As a result, prior knowledge that is deemed irrelevant in subsequent steps can be largely disregarded, leading to the enhancement of the quality of the reconstruction output. 

We then incorporate the adaptively filtered knowledge $\bm{l}_{\rm A}$ with the current input information $\bm{c}$ through a transformer layer:
\begin{equation}
    \bm{l}_{\rm Tran} = {\rm Transformer}(\bm{c}, \bm{l}_{\rm A}, \bm{l}_{\rm A})
    \label{equ:ras_transformer}
\end{equation}

Finally, the updated latent feature $\bm{l}^*$ can be derived by fusing the previous latent feature $\bm{l}$ and $\bm{l}_{\rm Tran}$:
\begin{equation}
    \bm{l}^* = (\bm{W}\bm{l} + \bm{l}_{\rm Tran}) / 2\\
\end{equation}

In order to restore it to the dimension of the original feature, we use a linear projection to get the output of the RASFormer block, which can be derived as follows:
\begin{equation}
    \bm{o} = \bm{W}_o\bm{l}^*
\end{equation}

Summing it up, the RASFormer block can be summarized into a function: 
\begin{equation}
    \bm{l}^*, \bm{o} = {\rm RASFormer}(\bm{c}, \bm{l})
\end{equation}

\textbf{Remarks.} The employed adaptive gate (\textit{i.e.,} Eq.~\ref{equ:reset}) can filter out the previously reconstructed information, preventing wastage of the decoder's reconstruction capacity. Also, it enables the decoder to fully consider the discrepancy between the previously reconstructed information and the currently to-be-reconstructed information, thereby achieving an enhancement of \textit{sequential dynamics} during the reconstruction process. Furthermore, thanks to the MHSA in the transformer layer (\textit{i.e.,} Eq.~\ref{equ:ras_transformer}), the RASFormer block can facilitate the effective interaction between each element in $\bm{c}$ and other elements in $\bm{l}$, enabling the capture of \textit{spatial dynamics}.

\subsection{Loss and Inference}
\label{sec:loss}
\textbf{Objective function.} The objective function for training RAS is to calculate the MSE loss between the original feature $\bm{f}_{\rm org}$ and the reconstructed feature $\bm{f}_{\rm rec}$.
\begin{align}
    \mathcal{L} = \frac{1}{H\times W}\Vert\bm{f}_{\rm org} - \bm{f}_{\rm rec}\Vert_2^2
\end{align}

\textbf{Inference.} During the inference phase, the feature-level anomaly map $\bm{S}_{\rm\, feat}$ is computed by measuring the L2 norm of the difference between $\bm{f}_{\rm org}$ and $\bm{f}_{\rm rec}$.
\begin{align}
\bm{S}_{\rm\, feat} = \Vert\bm{f}_{\rm org} - \bm{f}_{\rm rec}\Vert_2 \in \mathbb{R}^{H \times W}
\label{eq:inference}
\end{align}

The anomaly map is then up-sampled to the size of the original image using bi-linear interpolation to obtain the pixel-level anomaly map. The image-level anomaly score is derived by taking the maximum value of the averaged pooled pixel-level anomaly map.

%%%%%%%%%%%%%%%%%%%%%%%%%% Experiments %%%%%%%%%%%%%%%%%%%%%%%%%%%%%%%%%%

\begin{table*}[t]
\setlength\tabcolsep{4pt}
\caption{\text{Image-level AUROC for anomaly detection} on MVTec-AD (unified / \textcolor{gray}{separate}).}
\vspace{15pt}
\centering
\small
\begin{tabular}{c|cccccccc|l}
\toprule

Category &  US & PaDiM & CutPaste & MKD & DRAEM & SimpleNet & DeSTSeg & UniAD & \makecell[c]{RAS (ours)}\\
\midrule
Bottle  & 84.0 / \sep{99.0} & 97.9 / \sep{99.9} & 67.9 / \sep{98.2} & 98.7 / \sep{99.4} & 97.5 / \sep{99.2} & 98.7 / \sep{100} & \textbf{100} / \sep{100} & 99.7/ \sep{100} & \textbf{100} $\pm$ 0.00 / \sep{100} \\
Cable   & 60.0 / \sep{86.2} & 70.9 / \sep{92.7} & 69.2 / \sep{81.2} & 78.2 / \sep{89.2} & 57.8 / \sep{91.8} & 93.6 / \sep{99.9} & 94.5 / \sep{97.8} & 95.2/ \sep{97.6} & \textbf{99.2} $\pm$ 0.12 /  \sep{99.7}\\
Capsule    & 57.6 / \sep{86.1} & 73.4 / \sep{91.3} & 63.0 / \sep{98.2} & 68.3 / \sep{80.5} & 65.3 / \sep{98.5} & 73.7 / \sep{97.7} & 87.4 / \sep{97.0} & 86.9/ \sep{85.3} & \textbf{92.6} $\pm$ 0.32 /  \sep{95.6}\\
Carpet     & 86.6 / \sep{91.6} & 93.8 / \sep{99.8} & 93.6 / \sep{93.9} & 69.8 / \sep{79.3} & 98.0 / \sep{97.0} & 91.5 / \sep{99.7} & 98.1 / \sep{98.9} & \textbf{99.8} / \sep{99.9} & 99.5 $\pm$ 0.05 / \sep{100}\\
Grid       & 69.2 / \sep{81.0} & 73.9 / \sep{96.7} & 93.2 / \sep{100}  & 83.8 / \sep{78.0} & 99.3 / \sep{99.9} & 50.2 / \sep{99.7} & 98.4 / \sep{99.7} & 98.2/ \sep{98.5} & \textbf{99.8} $\pm$ 0.16 / \sep{100}\\
Hazelnut   & 95.8 / \sep{93.1} & 85.5 / \sep{92.0} & 80.9 / \sep{98.3} & 97.1 / \sep{98.4} & 93.7 / \sep{100}  & 98.1 / \sep{100} & 99.8 / \sep{99.9} & 99.8 / \sep{99.9} & \textbf{100} $\pm$ 0.00 / \sep{100}\\
Leather    & 97.2 / \sep{88.2} & 99.9 / \sep{100} & 93.4 / \sep{100}  & 93.6 / \sep{95.1} & 98.7 / \sep{100}  & 98.5 /\sep{100} & \textbf{100} / \sep{100} & \textbf{100} / \sep{100} & \textbf{100} $\pm$ 0.00 / \sep{100}\\
Metal Nut  & 62.7 / \sep{82.0} & 88.0 / \sep{98.7} & 60.0 / \sep{99.9} & 64.9 / \sep{73.6} & 72.8 / \sep{98.7} & 95.4 / \sep{100} & \textbf{100} / \sep{99.5} & 99.2 / \sep{99.0}& 99.9 $\pm$ 0.02 / \sep{99.4} \\
Pill       & 56.1 / \sep{87.9} & 68.8 / \sep{93.3} & 71.4 / \sep{94.9} & 79.7 / \sep{82.7} & 82.2 / \sep{98.9} & 87.9 / \sep{99.0} & 92.1 / \sep{97.2} & 93.7 / \sep{88.3} & \textbf{96.3} $\pm$ 0.35 / \sep{96.2}\\
Screw      &  66.9 / \sep{54.9} &56.9 / \sep{85.8} & 85.2 / \sep{88.7} & 75.6 / \sep{83.3} & 92.0 / \sep{93.9} & 65.1 / \sep{98.2} & 73.4 / \sep{93.6} & 87.5/ \sep{91.9}& \textbf{95.3} $\pm$ 0.40 / \sep{95.6} \\
Tile       & 93.7 / \sep{99.1} & 93.3 / \sep{98.1} & 88.6 / \sep{94.6} & 89.5 / \sep{91.6} & 99.8 / \sep{99.6} & 94.4 / \sep{99.8} & 99.3 / \sep{100} & 99.3/ \sep{99.0} & \textbf{100} $\pm$ 0.02 / \sep{99.9}\\
Toothbrush &  57.8 / \sep{95.3} &95.3 / \sep{96.1} & 63.9 / \sep{99.4} & 75.3 / \sep{92.2} & 90.6 / \sep{100}  & 85.3 / \sep{99.7} & 81.7 / \sep{99.9} & 94.2/ \sep{95.0} & \textbf{98.7} $\pm$ 0.30 / \sep{94.8}\\
Transistor & 61.0 / \sep{81.8} & 86.6 / \sep{97.4} & 57.9 / \sep{96.1} & 73.4 / \sep{85.6} & 74.8 / \sep{93.1} &  75.9 / \sep{100} & 95.0 / \sep{98.5} & \textbf{99.8}/ \sep{100} &  99.2 $\pm$ 0.00 / \sep{100}\\
Wood    & 90.6 / \sep{97.7} & 98.4 / \sep{99.2} & 80.4 / \sep{99.1} & 93.4 / \sep{94.3} & \textbf{99.8} / \sep{99.1} & 97.7 / \sep{100} & 100 / \sep{97.1} & 98.6/ \sep{97.9} & 98.7 $\pm$ 0.23 / \sep{98.5}\\
Zipper     & 78.6 / \sep{91.9} & 79.7 / \sep{90.3} & 93.5 / \sep{99.9} & 87.4 / \sep{93.2} & \textbf{98.8} / \sep{100}  & 97.8 / \sep{99.9} & 99.0 / \sep{100} & 95.8 / \sep{96.7} &  98.4 $\pm$ 0.07 \sep{99.4}\\
\midrule
Mean & 74.5 / \sep{87.7} & 84.2 / \sep{95.5} & 77.5 / \sep{96.1} & 81.9 / \sep{87.8} & 88.1 / \sep{98.0} & 86.9 / \sep{99.6} & 94.6 / \sep{98.6} & 96.5 / \sep{96.6} & \textbf{98.4} $\pm$ 0.08 / \sep{98.6}\\
\bottomrule
\end{tabular}

\label{tab:mvtec_image}
% \vspace{-15pt}
\end{table*}

\begin{table*}[ht]
\setlength\tabcolsep{4pt}
\caption{\text{Pixel-level AUROC for anomaly localization} on MVTec-AD (unified / \textcolor{gray}{separate}).}
\vspace{15pt}
\centering
\small
% \vspace{-0pt}
\begin{tabular}{c|cccccccc|l}
\toprule
Category & US & PaDiM & FCDD & MKD & DRAEM & SimpleNet & DeSTSeg & UniAD & \makecell[c]{RAS (ours)}\\
\midrule
Bottle     & 67.9 / \sep{97.8} & 96.1 / \sep{98.2} & 56.0 / \sep{97} & 91.8 / \sep{96.3} & 87.6 / \sep{99.1}  & 96.5 / \sep{98.0} & 98.2 / \sep{99.2} & 98.1 / \sep{98.1} & \textbf{98.4} $\pm$ 0.02 / \sep{98.5}\\
Cable      & 78.3 / \sep{91.9} & 81.0 / \sep{96.7} & 64.1 / \sep{90} & 89.3 / \sep{82.4} & 71.3 / \sep{94.7} & 91.1 / \sep{97.6} & 93.5 / \sep{97.3} & 97.3 / \sep{96.8} & \textbf{98.7} $\pm$ 0.03 / \sep{98.6}\\
Capsule    & 85.5 / \sep{96.8} & 96.9 / \sep{98.6} & 67.6 / \sep{93} & 88.3 / \sep{95.9} & 50.5 / \sep{94.3} & 92.2 / \sep{98.9} & 96.9 / \sep{99.1} & 98.5 / \sep{97.9} & \textbf{98.6} $\pm$ 0.01 / \sep{98.6}\\
Carpet  & 88.7 / \sep{93.5} & 97.6 / \sep{99.0} & 68.6 / \sep{96} & 95.5 / \sep{95.6} & \textbf{98.6} / \sep{95.5} & 96.0 / \sep{98.2}& 97.4 / \sep{96.1} & 98.5/ \sep{98.0} & 97.9 $\pm$ 0.07 / \sep{98.7}\\
Grid   & 64.5 / \sep{89.9} & 71.0 / \sep{97.1} & 65.8 / \sep{91} & 82.3 / \sep{91.8} & \textbf{98.7} / \sep{99.7} & 53.7 / \sep{98.8} & 96.6 / \sep{99.1} & 96.5 / \sep{94.6} & 97.1 $\pm$ 0.03 / \sep{97.2}\\
Hazelnut   & 93.7 / \sep{98.2} & 96.3 / \sep{98.1} & 79.3 / \sep{95} & 91.2 / \sep{94.6} & 96.9 / \sep{99.7} & 94.8 / \sep{97.9} & \textbf{99.0} / \sep{99.6} & 98.1 / \sep{98.8} & 98.5 $\pm$ 0.02 / \sep{98.7}\\
Leather& 95.4 / \sep{97.8} & 84.8 / \sep{99.0} & 66.3 / \sep{98} & 96.7 / \sep{98.1} & 97.3 / \sep{98.6} & 97.1 / \sep{99.2} & \textbf{99.6} / \sep{99.7} & 98.8 / \sep{98.3} & 98.7 $\pm$ 0.05 / \sep{99.2}\\
Metal Nut  & 76.6 / \sep{97.2} & 84.8 / \sep{97.3} & 57.5 / \sep{94} & 64.2 / \sep{86.4} & 62.2 / \sep{99.5} & 94.3 / \sep{98.8} & 97.0 / \sep{98.6} & 94.8 / \sep{95.7} & \textbf{97.3} $\pm$ 0.12 / \sep{98.1}\\
Pill       & 80.3 / \sep{96.5} & 87.7 / \sep{95.7} & 65.9 / \sep{81} & 69.7 / \sep{89.6} & 94.4 / \sep{97.6} & 92.5 / \sep{98.6}& 97.4 / \sep{98.7} &  95.0 / \sep{95.1} & \textbf{98.3} $\pm$ 0.11 / \sep{98.2}\\
Screw      & 90.8 / \sep{97.4} & 94.1 / \sep{98.4} & 67.2 / \sep{86} & 92.1 / \sep{96.0} & 95.5 / \sep{97.6} & 94.5 / \sep{99.3} & 94.6 / \sep{98.5} & 98.3 / \sep{97.4} & \textbf{99.1} $\pm$ 0.03 / \sep{99.1}\\
Tile    & 82.7 / \sep{92.5} & 80.5 / \sep{94.1} & 59.3 / \sep{91} & 85.3 / \sep{82.8} & \textbf{98.0} / \sep{99.2} & 90.9 / \sep{97.0} & 95.3 / \sep{98.0} & 91.8 / \sep{91.8} & 92.9 $\pm$ 0.14 / \sep{94.1}\\
Toothbrush & 86.9 / \sep{97.9} & 95.6 / \sep{98.8} & 60.8 / \sep{94} & 88.9 / \sep{96.1} & 97.7 / \sep{98.1} & 94.2 / \sep{98.5} & 97.7 / \sep{99.3} & \textbf{98.4} / \sep{97.8} & \textbf{98.4} $\pm$ 0.01 / \sep{98.5}\\
Transistor & 68.3 / \sep{73.7} & 92.3 / \sep{97.6} & 54.2 / \sep{88} & 71.7 / \sep{76.5} & 64.5 / \sep{90.9} & 84.5 / \sep{97.6} & 78.8 / \sep{89.1} & 97.9 / \sep{98.7} & \textbf{98.9} $\pm$ 0.03 / \sep{99.1}\\
Wood    & 83.3 / \sep{92.1} & 89.1 / \sep{94.1} & 53.3 / \sep{88} & 80.5 / \sep{84.8} & 96.0 / \sep{96.4} & 90.7 / \sep{94.5} & \textbf{97.9} / \sep{97.7} & 93.2 / \sep{93.4} & 92.0 $\pm$ 0.23 / \sep{92.9}\\
Zipper     & 84.2 / \sep{95.6} & 94.8 / \sep{98.4} & 63.0 / \sep{92} & 86.1 / \sep{93.9} & \textbf{98.3} / \sep{98.8} & 96.2 / \sep{98.9} & 98.0 / \sep{99.1} & 96.8 / \sep{96.0} & 97.8 $\pm$ 0.03 / \sep{97.7}\\

\midrule
Mean & 81.8 / \sep{93.9} & 89.5 / \sep{97.4} & 63.3 / \sep{92} & 84.9 / \sep{90.7} & 87.2 / \sep{97.3} & 90.6 / \sep{98.1} & 95.9 / \sep{97.9} & 96.8 / \sep{96.6} & \textbf{97.5} $\pm$ 0.01 / \sep{97.8}\\
\bottomrule
\end{tabular}

\label{tab:mvtec_pixel}
% \vspace{-5pt}
\end{table*}

\section{Experiments}
\subsection{Experiment Setups}

\textbf{Datasets.} We validate the effectiveness of the proposed RAS method by comparing it with several baseline methods on four widely used benchmark datasets for unsupervised anomaly detection, including MVTec-AD \cite{mvtec}, VisA \cite{visa}, BTAD \cite{btad}, and MPDD \cite{mpdd}. 

\begin{itemize}

\item MVTec-AD is a widely used benchmark for image anomaly detection, including 15 categories of industrial products and defects. It consists of 3,629 anomaly-free images for training and 1,725 images for testing. For the test set, both normal and anomalous samples are provided (467 normal images and 1258 anomalous images).

\item VisA comprises 12 subsets, each corresponding to a distinct object. 
It contains a total of 10,821 images, with 8,659 anomaly-free images in the training set. The test set consists of 2,162 images, including 962 normal and 1,200 anomalous images.

\item BTAD presents a real-world industrial anomaly dataset, consisting of a collection of 2,540 images capturing body and surface defects in three distinct industrial products.
The training set contains 1,799 normal images, and the test set includes 451 normal images and 290 anomalous images.

\item MPDD includes six types of metal parts and consists of 888 images in the training set. The test set comprises 176 normal images and 282 anomalous images. 

\end{itemize}

\textbf{Implementation details.}
We employ EfficientNet-B4 as the backbone. We resize the images to $224 \times 224$ before feeding them into the backbone. Feature maps are extracted from levels 1 to 4, resulting in a concatenated feature channel $C_{org}$ of 272. These features are aligned to the dimensions of the highest-level feature map, namely $14 \times 14$. In both the encoder and decoder, the channel dimension for the reconstructed latent feature $C_{rec}$ is set to 256. The Multi-Head Self-Attention (MHSA) uses 8 heads. We utilize the AdamW optimizer with a learning rate of $7e-4$ and a weight decay of $1e-4$. The batch size is set to 64. All models are trained with 500 epochs.

\textbf{Evaluation Metrics.} The performance of anomaly detection models is typically measured by AUROC. We report the image-level AUROC and the pixel-level AUROC on these four datasets, following previous work~\cite{uniad,us,draem}.

\subsection{Comparison with State-of-the-art Methods}

\textbf{Performance comparison on MVTec-AD}.
We select US~\cite{us}, PaDiM~\cite{padim}, MKD~\cite{mkd}, DRAEM~\cite{draem}, SimpleNet~\cite{simplenet}, DeSTSeg~\cite{destseg}, UniAD~\cite{uniad} as our baseline methods, representing various types of anomaly detection\footnote{Some of the latest methods, such as PNI \cite{pni} and OmniAL \cite{omnial}, have heavy time and space complexities in the unified unsupervised anomaly detection setting, requiring high computation and storage. We leave them out for fair comparison.}.
We compare our method with baselines under the two different paradigms mentioned in Fig \ref{fig:paradigm}, \textit{i.e.,} the \textit{unified} setting and the \textit{separate} setting. We report the performance at the image level and pixel level on MVTec-AD in Table \ref{tab:mvtec_image} and Table \ref{tab:mvtec_pixel}, respectively.
We can see that, for the \textit{unified} unsupervised anomaly detection, our method can outperform UniAD with a significant improvement of 1.9\% AUROC for image-level anomaly detection and of 0.7\% AUROC for pixel-level anomaly localization. Although our RAS is not specifically designed for the conventional \textit{separate} setting, it achieves comparable performance to conventional advanced methods. Compared to UniAD, our method can also obtain a 2.0\% improvement in image-level AUROC and a 1.2\% increase in terms of the pixel-level AUROC in the separate setting.

\textbf{Performance comparison on VisA, BTAD and MPDD}.
For these datasets, we select DRAEM~\cite{draem}, SimpleNet~\cite{simplenet}, and DeSTSeg~\cite{destseg} as baselines due to their remarkable performance under the unified setting. Apart from results of individual dataset, we also report their average as the overall performance.
Table \ref{tab:3_image_pixel} shows the comparison results.
We can observe that RAS presents superior performance to baseline methods on average.
Compared to DeSTSeg, our method can achieve an average improvement of 0.5\% and 1.4\% in image-level and pixel-level AUROC, respectively. 
Such performance gains across different datasets well demonstrate the effectiveness and superiority of our method.

\begin{table*}[!ht]
\setlength\tabcolsep{4pt}
\caption{Comparison under the unified setting on VisA, BTAD, and MPDD.}
\vspace{15pt}
\centering
% \small
% \vspace{-5pt}
\begin{tabular}{c|cc|cc|cc|cc|cc}
\toprule
\multirow{2}{*}{\diagbox{Dataset}{Method}}
 & \multicolumn{2}{c|}{DRAEM}  & \multicolumn{2}{c|}{SimpleNet} & \multicolumn{2}{c|}{DeSTSeg} & \multicolumn{2}{c|}{UniAD} & \multicolumn{2}{c}{RAS (ours)}\\
\cmidrule{2-11}
  & I-AUC & P-AUC & I-AUC & P-AUC & I-AUC & P-AUC & I-AUC & P-AUC & I-AUC & P-AUC\\
\midrule
VisA & 74.5 & 84.7  &  87.9 & 95.1  & 88.6 & 96.0  & 88.6 & 98.3  & \textbf{92.9} & \textbf{98.7} \\
BTAD & 90.6 & 92.4 & 93.4 & 96.2  & 93.9 & \textbf{97.2}  & 92.3 & 97.1  & \textbf{94.7} & 97.0 \\
MPDD & 86.9 & 90.6  & 92.5 & 96.3  & \textbf{95.6} & 95.8  & 87.5 & 95.6  & 92.1 & \textbf{97.5} \\
\midrule
Avg  & 84.0 & 89.2  & 91.3 & 95.9  & 92.7 & 96.3  & 89.5 & 97.0  & \textbf{93.2} & \textbf{97.7} \\
\bottomrule
\end{tabular}
\vspace{-3pt}
\label{tab:3_image_pixel}
\vspace{5pt}
\end{table*}

\begin{table*}[ht]
\begin{minipage}{.5\linewidth}
\caption{Ablation study of the adaptive gate and the transformer.}
\vspace{15pt}
\centering
    % \vspace{-5pt}
    % \small
    \begin{tabular}{cc|cc}
        \toprule
        adaptive gate & transformer & I-AUROC & P-AUROC \\
        \midrule
        \checkmark &   & 97.5 & 97.3 \\
         & \checkmark & 96.1 & 97.2 \\
        \checkmark & \checkmark & \textbf{98.4} & \textbf{97.5} \\
        \bottomrule
    \end{tabular}
    \vspace{-3pt}
    
    \label{tab:ablation}
\end{minipage}%
\begin{minipage}{.5\linewidth}
\caption{Impact of the number of encoder-decoder (I-AUC / P-AUC).}
\vspace{15pt}
\centering
    % \small
    % \vspace{-5pt}
    \begin{tabular}{c|c|c|c|c}
        \toprule
        & $T_{d}$ = 1 & $T_{d}$ = 2 & $T_{d}$ = 3 & $T_{d}$ = 4 \\
        \midrule
        $T_{e}$ = 0 & 96.5 / 97.2 & 97.7 / 97.3 & 97.9 / 97.3 & 98.0 / 97.4\\
        $T_{e}$ = 1 & 97.1 / 97.3 & 97.8 / 97.4 & 98.2 / 97.5 & 98.3 / 97.4\\
        $T_{e}$ = 2 & 97.6 / 97.4 & 98.0 / 97.5 & 98.2 / 97.5 & \textbf{98.4 }/ \textbf{97.5}\\
        \bottomrule
    \end{tabular}
    \vspace{-3pt}
    
    \label{tab:num_ed}
\end{minipage}
% \vspace{-5pt}
\end{table*}

\begin{table*}[!ht]
\caption{Impact of noise during the feature reconstruction. Values in parentheses indicate differences from results of noise intensity of 0.}
\vspace{15pt}
\centering
% \small
\begin{tabular}{c|c|c|ccccc}
\toprule
 \multicolumn{2}{c|}{noise intensity, \textit{i.e.,}  $\alpha$ } & 0 & 10 & 20 & 30 & 40 & 50 \\
\midrule
\multirow{2}{*}{image-level}
& baseline & 96.5 & 96.4 \text{(-0.1)} & 96.2 \text{(-0.3)} & 95.1 \text{(-1.4)} & 90.9 \text{(-5.6)} & 83.7 \text{(-12.8)} \\
& RAS   & \textbf{98.4} & 98.4 \textbf{(-0.0)} & 98.2 \textbf{(-0.2)} & 97.6 \textbf{(-0.8)} & 96.6 \textbf{(-1.8)} & 91.9 \textbf{(-6.5)} \\
\midrule
\multirow{2}{*}{pixel-level}
& baseline & 96.8 & 96.8 \text{(-0.0)} & 96.7 \text{(-0.1)} & 96.5 \text{(-0.3)} & 95.5 \text{(-1.3)} & 92.0 \text{(-4.8)} \\
& RAS   & \textbf{97.5} & 97.5 \textbf{(-0.0)} & 97.5 \textbf{(-0.0)} & 97.4 \textbf{(-0.1)} & 97.2 \textbf{(-0.3)} & 96.4 \textbf{(-1.1)} \\
\bottomrule
\end{tabular}
\label{tab:noise}
\end{table*}

\subsection{Model Analysis}
\textbf{Ablation study.}
We analyze the impact of the adaptive gating strategy and the transformer in the RASFormer block. As shown in Table \ref{tab:ablation}, with only the adaptive gate, the performance will reduce by 0.9\% and 0.2\% in terms of the image-level AUROC and the pixel-level AUROC, respectively. Using only the transformer can greatly decrease the performance to 96.1\%/97.2\%, which is 2.3\%/0.3\% behind our RAS (the fourth row in Table \ref{tab:ablation}). These results indicate the positive effect of the proposed adaptive gating strategy and the involvement of the transformer, which can be attributed to the benefit of capturing the spatial dynamics and the sequential dynamics using them in our RAS.

\textbf{Analysis of the number of encoder-decoder layers.}
We investigate the performance of RAS with different numbers of encoder-decoder layers in Table \ref{tab:num_ed}.
We can observe that increasing the number of encoder layers or decoder layers can bring a substantial performance improvement. Surprisingly, feeding CNN features directly into one RASFormer decoder without the encoder, \textit{i.e.,} $T_e=0, T_d=1$, can still yield quite satisfactory results (96.5\%/97.2\%). 
Adding one more decoder layer \textit{i.e.,} $T_e=0, T_d=2$, can result in performance improvements, especially for the image-level AUROC (1.2\%).
We can also observe that compared to one decoder, more decoders can lead to consistent performance improvement, indicating the remarkable benefit of modeling the feature reconstruction from the sequence perspective.

\begin{figure*}[hb]
\centering
\includegraphics[width=0.85\linewidth]{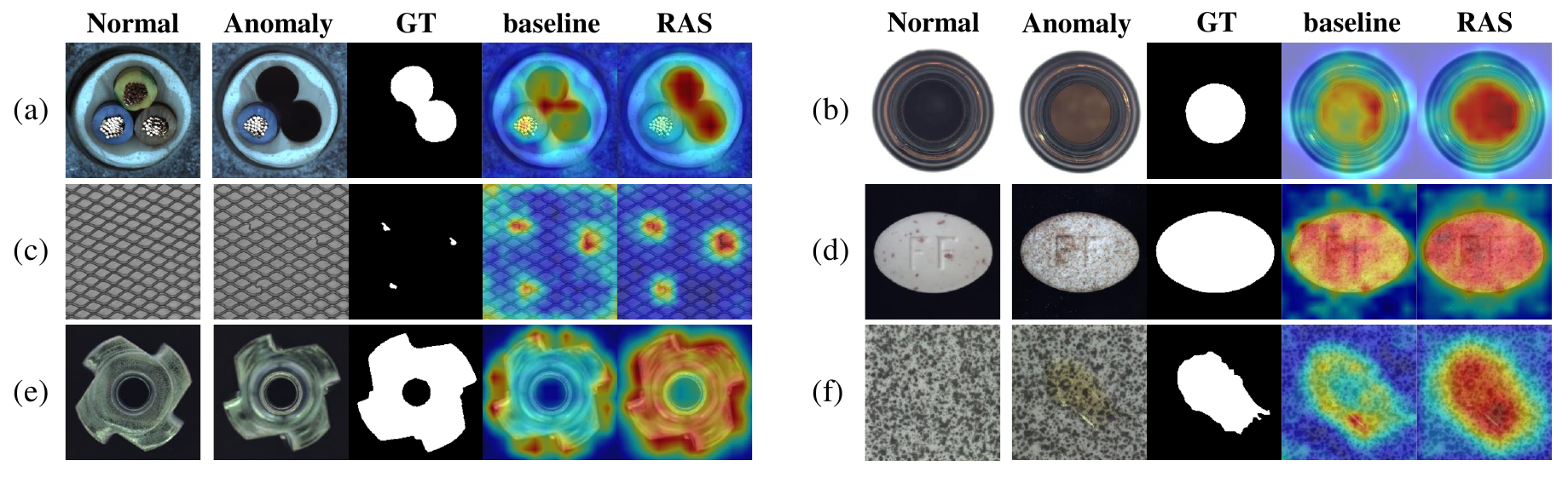}
\vspace{0pt}
\caption{\textbf{Qualitative results for anomaly map} on MVTec-AD. We turn the anomaly map into the heat map for better visualization. Regions with higher anomaly scores are depicted in vibrant red colors. Best viewed in colors. ``GT'' means the ground truth.} 
\label{fig:anomaly_map_mvtec}
\end{figure*}

\textbf{Analysis of noise during the reconstruction}. Intuitively, the noise information, \textit{i.e.,} $\bm{\epsilon}$ in Eq. \ref{eq:noise} serves as a simulation of the anomaly distribution during the feature reconstruction process. During training, our RAS can be seen as a mapping function that maps various features to the normal distribution, regardless of whether they are considered normal or abnormal. As a result, during testing ($\bm{\epsilon}$ is not applied), abnormal regions can be highlighted through the difference derivation in Eq.~\ref{eq:inference}. Therefore, it is worth investigating the impact of noise intensity for model inference, \textit{i.e.,} $\alpha$ in Eq. \ref{eq:noise}, to gain further insights into the effectiveness of our RAS. Here, we introduce UniAD as the baseline because of its advanced performance in the unified setting on the MVTec-AD dataset. We aim to evaluate the robustness of \texttt{well-trained} models. Therefore, during testing, the noise intensity, i.e., $\alpha$ is adjusted to assess the noise tolerance of pre-trained models. As shown in Table \ref{tab:noise}, when $\alpha$ = 0, no noise is introduced for model inference, thereby achieving the best performance for both baseline and RAS. We observe that when $\alpha=10$, RAS performs nearly on par with the condition of $\alpha=0$, while the baseline experiences a slight 0.1\% decrease in image-level AUROC. As we increase the noise scale from 20, the performance gap between the two models gradually widens. Notably, under intense noise with $\alpha=50$, the performance of baseline method significantly degrades, exceeding that of RAS by more than 4.0$\times$ in image-level AUROC and approximately 2.0$\times$ in pixel-level AUROC. Our method is more resistant to noise interference, exhibiting better robustness than baseline. This evidence clearly showcases the benefits of context enhancement by our RAS, effectively demonstrating the robustness, effectiveness, and superior ability of RAS in handling practical anomalies.

\begin{figure*}[!ht]
\centering
\includegraphics[width=0.9\linewidth]{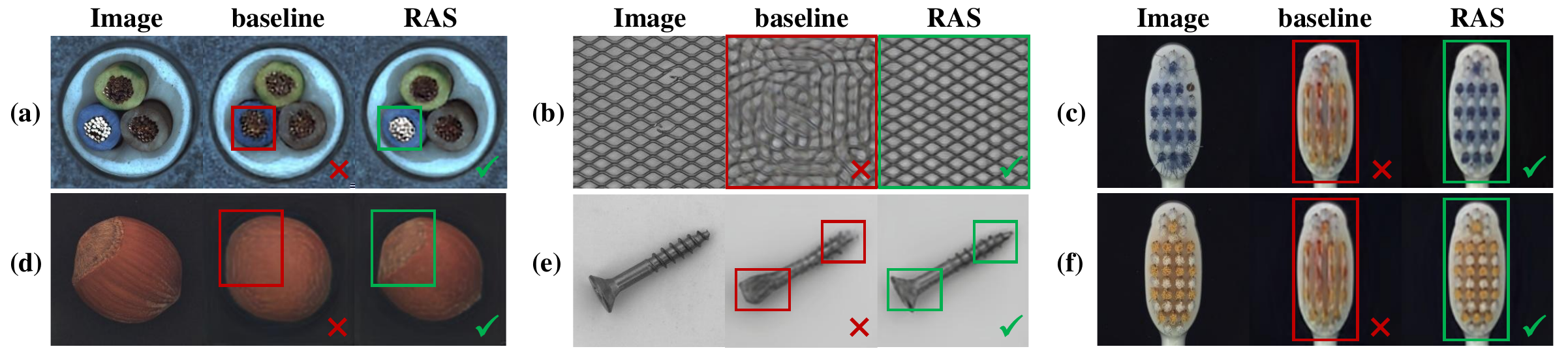}
\vspace{3pt}
\caption{\textbf{Visualization comparison of image reconstruction}. We utilize bounding boxes to visually differentiate between the worse (red) and better (green) regions.} 
\label{fig:image_recon_mvtec}
\end{figure*}

\begin{figure*}[!ht]
\centering
    \includegraphics[width=\textwidth]{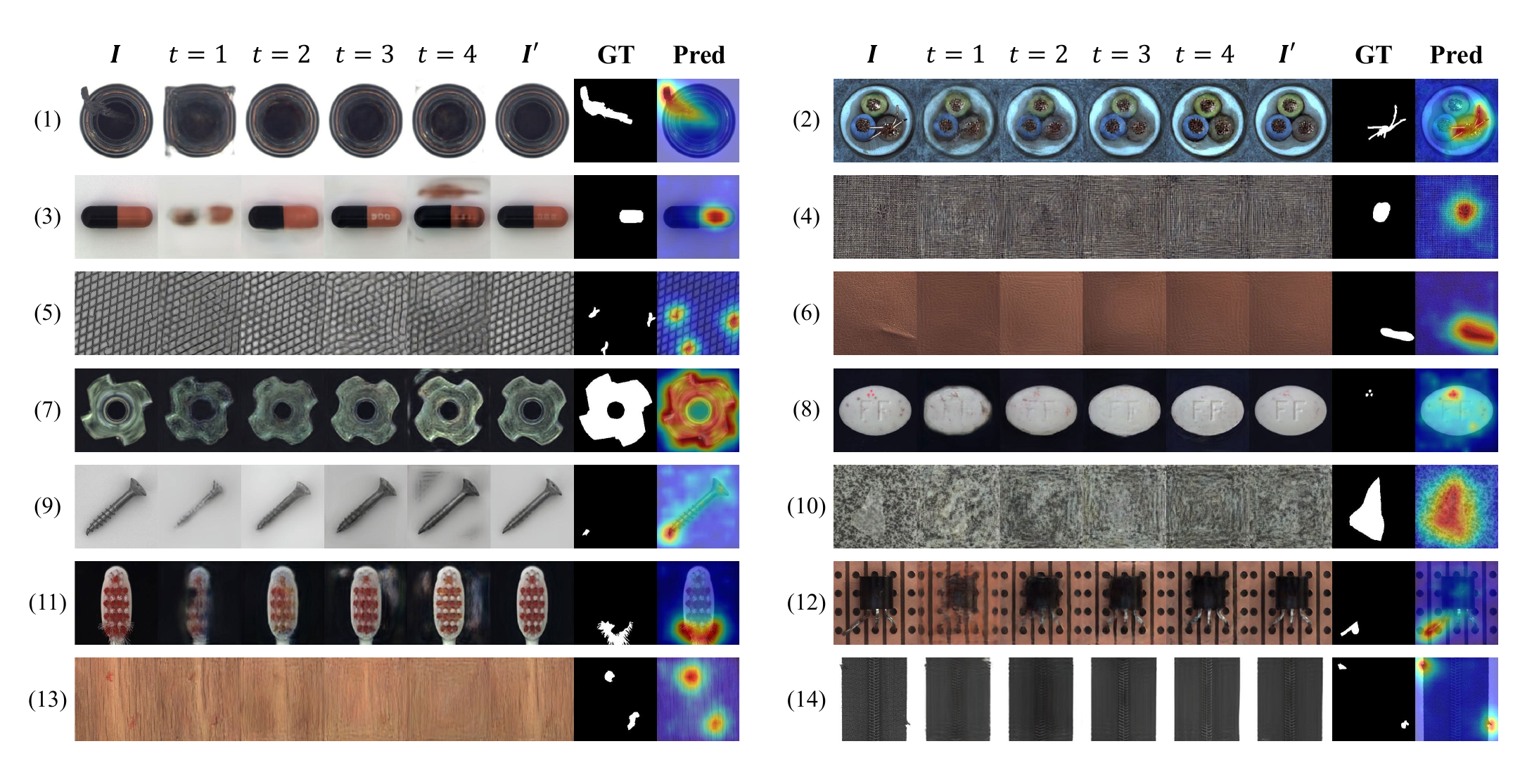}
    \vspace{-15pt}
    \caption{Visualization of the reconstruction process. $\bm{I}$ represents the original image, and t=1 to t=4 are images corresponding to the reconstructed features from each RASFormer decoder layer. $\bm{I'}$ is the final reconstructed image, ``\textbf{GT}'' is the binary mask, and ``\textbf{Pred}'' is the anomaly map predicted by our model.}
    \label{fig:anomaly}
    \vspace{8pt}
\end{figure*}

\subsection{Qualitative Results}
\textbf{Visualization of anomaly map.} To intuitively reveal the advantage of our proposed RAS model, we conduct a qualitative investigation of the anomaly maps generated by UniAD as a baseline and our RAS. As shown in Fig. \ref{fig:anomaly_map_mvtec}, it is evident that our RAS can localize the anomaly regions more accurately. For instance, in examples (a), (c), and (e), our proposed method generates more accurate anomaly maps compared to the baseline. Moreover, in examples (b), (d), and (f), RAS successfully emphasizes the salience of anomalous regions by yielding higher anomaly scores. These qualitative findings effectively demonstrate the benefits of enhancing the contextual awareness capability during feature reconstruction, highlighting the superiority of our RAS.

\textbf{Quality of image reconstruction.} The superiority of our method is not only evident in the anomaly maps but also reflected in the detailed image reconstruction. Fig. \ref{fig:image_recon_mvtec} presents a side-by-side comparison of the reconstructed images generated by RAS and UniAD. It is clearly observed that RAS provides a more accurate reconstruction of image details. For example, in (a), RAS accurately reproduces the reflection of the cable wire in the left-bottom area. In (e), RAS correctly replicates the head and tail of the screw, while UniAD fails. These results demonstrate that the reconstructed features in RAS are more aligned with the ground truth, resulting in superior image reconstruction and anomaly detection performance.

\textbf{Visualization of the reconstruction process.} To better illustrate the effectiveness of the contextual awareness capability, we also visualize the images corresponding to the reconstructed features from each RASFormer decoder layer. As shown in Fig. \ref{fig:anomaly}, we can see that the reconstructed images effectively repair the areas where defects are present, resulting in accurate anomaly maps compared to the original images. Cases (5) and (6) demonstrate that our model can reconstruct complex textures of carpets and grids. In the case of (12), where the backgrounds are intricate, our method remains unaffected and accurately identifies anomalies in the positions of transistor pins. 

By examining the features reconstructed by consecutive decoders \textit{i.e.}, from $t=1$ to $t=4$, we can also observe that the reconstruction process in our RAS roughly follows a coarse-to-fine pattern. 
As a result, the output of each decoder shows a significant improvement compared to the previous time step.
These results indicate that the proposed RAS can well perceive previously reconstructed information and then progressively calibrate the decoding outcome as the reconstruction process proceeds.

%%%%%%%%%%%%%%%%%%%%%%%%%% Conclusion %%%%%%%%%%%%%%%%%%%%%%%%%%%%%%%%%%
\section{Conclusion}
In this paper, we propose a novel Reconstruction as Sequence (RAS) framework for unified unsupervised anomaly detection. The main goal of our RAS is to enhance the contextual correspondence among different steps of feature reconstruction. To this end, we rethink the feature reconstruction from the sequence perspective with a generic RASFormer block. Inside the proposed RASFormer block, we adapt the transformer architecture with a novel strategy of adaptive gating. Thanks to the RASFormer block, our RAS can enhance the contextual awareness capability during feature reconstruction, leading to superior performance. Experimental results on standard benchmark datasets show that the proposed RAS can consistently outperform competing methods by a notable margin. These results well demonstrate the effectiveness and superiority of the proposed method.

\begin{ack}
This work was supported by National Science and Technology Major (No. 2022ZD0119401), National Natural Science Foundation of China (No. 61925107), the Key R \& D Program of Xinjiang, China (No. 2022B01006), and Zhejiang Provincial Natural Science Foundation of China under Grant (No. LDT23F01013F01).
\end{ack}

% \bibliography{mybibfile}
\bibliography{ecai-sample-and-instructions}

\end{document}